\documentclass[a4paper,twoside]{article}

\usepackage{epsfig}
\usepackage{subfig}
\usepackage{calc}
\usepackage{amssymb}
\usepackage{amstext}
\usepackage{amsmath}
\usepackage{amsthm}
\usepackage{multicol}
\usepackage{pslatex}
\usepackage{apalike}
\usepackage{algorithm2e}
\usepackage[bottom]{footmisc}
\usepackage{SCITEPRESS}
\usepackage[hidelinks]{hyperref}
\DeclareMathOperator*{\argmax}{arg\,max\,}
\DeclareMathOperator*{\argmin}{arg\,min\,}
\usepackage[export]{adjustbox}
\usepackage{enumerate}
\usepackage{mathrsfs}
\usepackage{booktabs}
\usepackage{tabularx}

\begin{document}

\title{ClusterComm: Discrete Communication in Decentralized MARL using Internal Representation Clustering}

\author{\authorname{Robert Müller\sup{1}, Hasan Turalic\sup{1}, Thomy Phan\sup{1}, Michael Kölle\sup{1}, Jonas Nüßlein\sup{1} and Claudia Linnhoff-Popien}
\affiliation{\sup{1}Institute of Informatics, LMU Munich, Munich, Germany}
\email{robert.mueller@ifi.lmu.de}
}

\keywords{Multi-Agent Reinforcement Learning, Communication, Clustering}

\abstract{
In the realm of Multi-Agent Reinforcement Learning (MARL), prevailing approaches exhibit shortcomings in aligning with human learning, robustness, and scalability. Addressing this, we introduce ClusterComm, a fully decentralized MARL framework where agents communicate discretely without a central control unit. ClusterComm utilizes Mini-Batch-K-Means clustering on the last hidden layer's activations of an agent's policy network, translating them into discrete messages. This approach outperforms no communication and competes favorably with unbounded, continuous communication and hence poses a simple yet effective strategy for enhancing collaborative task-solving in MARL.
}

\onecolumn \maketitle \normalsize \setcounter{footnote}{0} \vfill

\section{INTRODUCTION} \label{sec:introduction}
Problems such as autonomous driving~\cite{Shalev2016} or the safe coordination of robots in warehouses~\cite{salzman2020research} require cooperation between agents. In order to solve tasks collaboratively, agents must learn to coordinate by sharing resources and information. Similar to humans, cooperation can be facilitated through communication. The ability to communicate allows autonomous agents to exchange information about their current perceptions, their planned actions or their internal state. This enables them to cooperate on tasks that are more difficult or impossible to solve in isolation and without communication.\\
In Multi-Agent Reinforcement Learning~(MARL), multiple agents interact within a shared environment. In the simplest case, each agent is modeled as an independently learning entity that merely observes the other agents (Independent Learning). In this scenario, however, at best implicit communication can take place, as for example with bees that inform their peers about the position of a found food source by certain movements~\cite{von1992decoding}.
In the context of MARL, it has already been established that cooperation can be improved through the exchange of information, experience and the knowledge of the individual agents~\cite{tan_multi-agent_1993}.\\
The predominant approach to this is to provide the agents with additional information during training, such as the global state of the environment at the current timestep.
It is usually desirable to be able to execute the agents in a decentralized manner after training, i.e. each agent should only be able to decide how to act on the basis of information that is available to it locally. Consequently, the additional information at training time only serves to speed up training and to develop more sophisticated implicit coordination and communication strategies. 
This paradigm is referred to as Centralized Learning and Decentralized Execution.\\
An even more radical approach is to view all agents as a single RL agent that receives all the local observations of the individual agents and model the joint action space as a Cartesian product of the individual action spaces. Since the joint action space grows exponentially with the number of agents and agents cannot be deployed in a decentralized fashion, this approach is rarely used in practice.\\
To enable agents to communicate with each other explicitly, recent approaches~\cite{foerster2016learning,sukhbaatar2016learning,lowe2017multi,das2019tarmac,lin2021learning,wang2022fcmnet} introduce a communication channel between agents.
This approach poses a considerable exploration problem, as the probability of finding a consistent communication protocol is extremely low~\cite{foerster2016learning}. Furthermore, the initially random messages add to the already high variance of MARL and further complicate training~\cite{lowe2017multi,eccles2019biases}. Many works therefore resort to differentiable communication~\cite{sukhbaatar2016learning,lowe2017multi,choi2018compositional,mordatch2018emergence,eccles2019biases,wang2022fcmnet}, where agents can
can directly optimize the communication strategies of the other agents through the exchange of gradients. To further ease training, some approaches assume the messages the agents exchange to be continuous, yielding a high-bandwidth communication channel and thereby introducing scalability and efficiency issues.\\
Neither can humans adapt the neurons of their counterpart, nor is it always possible to gather additional information about the overall state of the environment while learning.
At its core, human communication relies upon discrete symbols, i.e. words. Discrete communication may offer greater practical utility (e.g., low-bandwidth communication channels between robots~\cite{lin2021learning}).\\
Besides, it is common practice in MARL to use the same neural network for all agents (\textit{parameter sharing}). This allows the network to benefit from the entire experience of all agents and speeds up training. Again, this assumption does not translate well to how human individuals gather knowledge. Moreover, it is reasonable to assume that the joint training of agents in the real world (without a simulation environment) should require no central control unit to avoid scalability issues and to increase robustness to malicious attacks~\cite{zhang2018fully}.\\
Based on the above, in this work we assume that each agent has its own neural network, that there exists a non-differentiable discrete communication channel, that there is no central control unit monitoring and influencing the training and that the agents can receive any additional information solely through the communication channel.\\
To this end, we propose \textit{ClusterComm} to enable discrete communication in fully decentralized MARL with little performance loss compared to the unbounded approach. The internal representation (policy's penultimate layer) of an agent is used directly to create a message. The internal representation. By clustering the representations using K-Means, a discrete message can subsequently be created using the index of the assigned cluster. 
\section{BACKGROUND} \label{sec:preliminaries}
In this section, we outline the concepts this work builds upon. Markov games are the underlying formalism for MARL and K-Means is the clustering algorithm ClusterComm uses to discretize messages.

\subsection{Markov-Games}
In this work, we consider Partially Observable Markov Games (POMGs)~\cite{shapley1953stochastic,littman1994markov,hansen2004dynamic}. At each time step, each agent receives a partial observation of the full state. These observations are in turn used to learn appropriate policies that maximize the individual reward. \\ Formally, a POMG with $N$ agents is defined by the tuple 
    $(\mathscr{N}, \mathcal{S}, \mathcal{A}, \mathcal{O}, \Omega, \mathcal{P}, \mathcal{R}, \gamma)$ with $\mathscr{N} =\{1, \dots, N\}$. Furthermore, $\mathcal{S}$ denotes the set of states, $\mathcal{A}=\{\mathcal{A}^{(1)}, \dots, \mathcal{A}^{(N)}\}$ the set of actions and $\mathcal{O}=\{\mathcal{O}^{(1)}, \dots, \mathcal{O}^{(N)}\}$ the set of observation spaces.\\
At time $t$, agent $k \in \mathscr{N}$ observes a partial view $o_t^{(k)} \in \mathcal{O}^{(k)}$ of the underlying global state $s_t \in \mathcal{S}$ and subsequently selects the next action $a_t^{(k)} \in \mathcal{A}^{(k)}$. The agents' stochastic observations $o_t^{1:N} = \{o_t^{(1)}, \dots, o_t^{(N)}\}$ thus depend on the actions taken by all agents $a_{t-1}^{1:N}$ as well as the global state $s_t$ via $\Omega \left(o_{t}^{1:N} \vert s_t, a_{t-1}^{1:N} \right)$. Similarly, the stochastic state $s_t$ depends on the previous state and the chosen actions of all agents $\mathcal{P}\left(s_t \, \vert \, s_{t-1}, a_{t-1}^{1:N} \right)$. Finally, each agent $k$ receives a reward according to an individual reward function $\mathcal{R}^{(k)}: \mathcal{S} \times \mathcal{A} \times \mathcal{S} \rightarrow \mathbb{R},\,\, \mathcal{R} = \{R^{(1)}, \dots, R^{(N)}\}$. Let $r_t^{(k)}$ be the reward received by agent $k$ at time $t$ and the agent's policy $\pi^{(k)}\left(a_t^{(k)} \vert o_t^{(k)}\right)$ a distribution over actions to be chosen given the local observation.
The goal is to find policies $\pi^{1:N}$ such that the discounted reward
\begin{equation}
    G^{(k)} = \sum_{t=0}^{T} \gamma^t r_t^{(k)}
\end{equation}
is maximized with respect to the other policies in $\pi$:
\begin{equation}
    \forall_k : \pi^{(k)} \in \argmax_{\Tilde{\pi}^{(k)}} \mathbb{E}[G^{(k)} \, \vert \, \Tilde{\pi}^{(k)},\pi^{(-k)}]
\end{equation}
where $\pi^{(-k)} = \pi \setminus \{\pi^{(k)}\}$.\\
To explicitly account for the possibility of communication in a POMG~\cite{lin2021learning}, the POMG can be extended to include the set of message spaces $\mathcal{M} = \{\mathcal{M}^{(1)}, \dots, \mathcal{M}^{(N)}\}$. At time $t$, each agent now has access to all messages from the previous time step $m_t = \{m_{t-1}^{(k)} \in \mathcal{M}^{(k)}\, \vert \, 1 \leq k \leq N\}$. The policy then also processes and receives new messages $\pi^{(k)}\left(a_t^{(k)}, m_t^{(k)} \vert o_t^{(k)},m_{t-1}^{(k)}\right)$.

\subsection{K-Means}
\label{sec:preliminaries:kmeans}
Clustering is the most common form of unsupervised learning. Here, data is divided into groups based on their similarity without having to assign any labels beforehand. The goal is to find a clustering in which the similarity between individual objects in one group is greater than between objects in other groups. To determine the similarity, distance or similarity functions such as Euclidean distance or cosine similarity are typically used.\\
K-Means~\cite{lloyd1982least} is one of the simplest and most widely used clustering algorithms and is adopted in this work to discretize the agent's internal representations.\\
Let $D = \{x_i \in \mathbb{R}^D \, \vert ,\ 2 \leq i \leq n\}$ be a dataset of $n$ vectors with $D$ dimensions and $\mu = \{\mu_i \in \mathbb{R}^D \vert 2 \leq i \leq k\}$ be a set of $k \geq 2$ centroids of the same dimensionality. 
Further, let $\mu_{x_i} := \argmin_{\mu_j \in \mu} \Vert x_i - \mu_j\Vert_2^2$ be the closest centroid of $x_i$.
The the objective is to to divide $D$ into $k$ partitions (quantization) such that the sum of the squared euclidean distances from the centroids is minimal, i.e.:
\begin{equation}
    \argmin_{\mu} \sum_{i=1}^n \Vert x_i - \mu_{x_i} \Vert_2^2
\end{equation}
Since the search for the optimal solution is NP-hard, one resorts to approximate algorithms such as Lloyds algorithm~\cite{lloyd1982least}. Here, the centroids are initially chosen at random and subsequently each data point is assigned to the centroid to which the distance is minimal. The new centroids are then set to the mean value of all data points assigned to the respective centroid. The last two steps are repeated until convergence.\\
Mini-Batch-K-Means~\cite{sculley2010web} is designed to reduce the memory consumption and allows for iterative re-clustering. It considers only a fixed-size subset of $D$ in each iteration (mini-batches). In addition, Mini-Batch-K-Means is useful in scenarios where the entire data set is unknown at any given time and may be subject to time-dependent changes (e.g. data streams, concept-drift). This property is of particular importance for ClusterComm since the underlying featurespace exhibits non-stationarity during training.
\section{CLUSTER COMM} \label{sec:approach}
ClusterComm is loosely inspired by the way humans have learned to communicate over time. Its evolution started from simple hand movements and imitation of natural sounds to spoken language and is now an established communication protocol~\cite{fitch_evolution_2010}. Because human communication emerged through the interaction of individuals without supervision or a central control mechanism, ClusterComm eschews parameter sharing and differentiable communication, unlike most approaches presented in Section~\ref{sec:introduction}.
Instead, all agents learn and act independently, while still having the ability to exchange discrete messages through a communication channel.\\
With the aim of creating a communication mechanism that uses as few assumptions as possible, but as many as necessary, we propose to discretize the internal representation of each agent (last hidden layer of policy net) using Mini-Batch K-Means and use the resulting cluster indices as message. A message is thus made up of a single integer. Doing so, gives the agents access to a compact description of the observations and intended actions of all other agents, information that is often sufficient to solve a task collaboratively.\\
Clustering allows agents to compress information by grouping data based on similarities. In this way, agents are able to generate meaningful yet compact messages.\\
Discrete communication is especially difficult because without additional mitigations, no gradients can flow through the communication channel~\cite{vanneste2022analysis}. However, it is more in line with human communication as it evolved from sounds and gestures to symbols and words that are discrete~\cite{tallerman2005language}.

\begin{figure}[ht]
  \centering
  \includegraphics[width=1.0\linewidth]{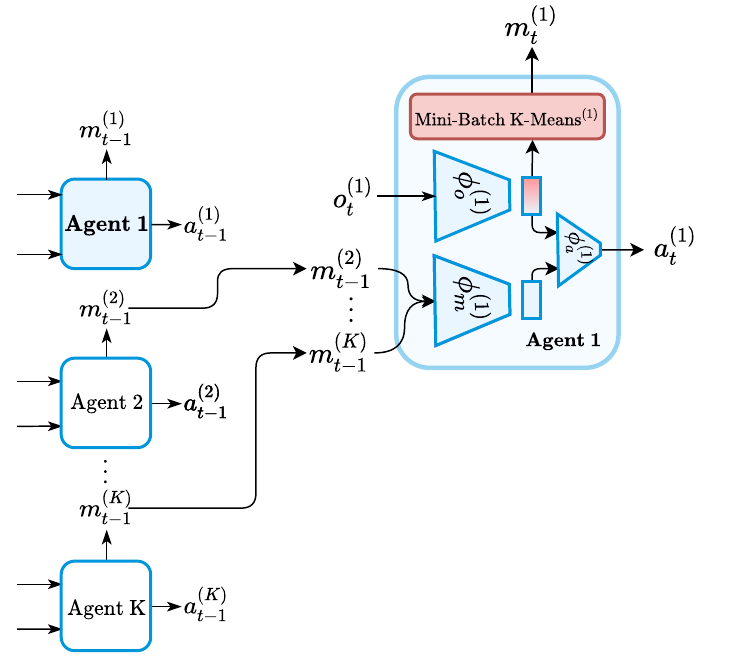}
  \caption{A visual depiction of ClusterComm's workflow for agent $1$. 
  At time $t$, agent $1$ receives observation $o_t^{(1)}$ and the messages $m_{t-1}^{(2)}, \dots, m_{t-1}^{(n)}$ from the other $n-1$ agents. The output of the message encoder $\phi_m^{(1)}\left(m_{t-1}^{2:N}\right)$ and observation encoder $\phi_o^{(1)}\left(o_t^{(1)}\right)$ is concatenated and passed through $\phi_{a}^{(1)}$ to compute the next action $a_t^{(1)}$ and the next message $m_t^{(1)}$. Subsequently, messages are discretized using $\text{Mini-Batch K-Means}^{(1)}$ by choosing the cluster index of the closest centroid.}
  \label{fig:clustercomm}
\end{figure}

\subsection{Algorithm}
At each time step $t$, each agent $k \in \{1, \dots, N\}$ receives two inputs: the current local observation $o_t^{(k)}$ and the messages $m_{t-1}^{(-k)} = \{m_{t-1}^{(i)} \vert 1 \leq i \leq n \land i \neq k\}$ sent by all other agents. Figure \ref{fig:clustercomm} illustrates the communication between agents for a single time step. Each agent uses a neural \textit{observation encoder} $\phi_o^{(k)}$ to create a representation for $o_t^{(k)}$. The \textit{message encoder} $\phi_m^{(k)}$ receives the concatenation of $m_{t-1}^{(-k)}$ (as one-hot encoding) and also produces a representation of the messages received. Finally, $\phi_a^{(k)}$ receives the concatenation of the representations of the messages and the local observation and computes the probability distribution over the action space $\mathcal{A}^{(k)}$.
Since $\phi_o^{(k)}\left(o_t^{(k)}\right)$ encodes information about the agent's observation and intended action, the output is discretized using Mini-Batch K-Means. The cluster index of the centroid closest to the output of the observation encoder is chosen as the message.
Since the parameters of the agents' networks are continuously adjusted during training (via gradient descent), the spanned feature space also varies over the course of training. The naïve application of K-Means would lead to the centroids being recalculated in each training step and consequently the indices being reassigned.
As a result, the messages sent would lose their meaning due to the constantly changing mapping from features to indices. The continuously evolving feature space resembles a data stream and thus the use of Mini-Batch K-Means can alleviate the aforementioned issues. The centroids are continuously adjusted with the occurrence of new data points, thereby changing the assignment less frequently.\\
To train tge agents, we use Proximal Policy Optimisation (PPO)~\cite{schulman2017proximal}. PPO is one of the most widely used algorithms in RL and is known for being very robust to the choice of hyperparameters. Moreover, its superiority in the context of MARL has been confirmed in recent studies~\cite{yu2021surprising,papoudakis2021benchmarking}.

\begin{figure*}[htb]
    \centering
    \subfloat[\centering][RedBlueDoors]{{
    \includegraphics[width=0.2\textwidth,trim=0 0 0 0, valign=c]{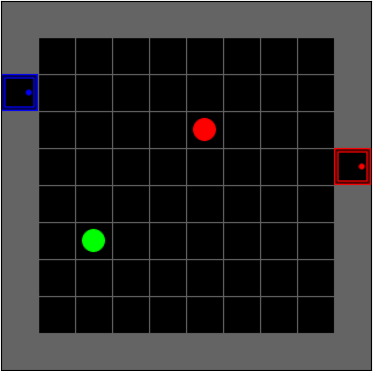}
    }}%
    \hfill
    \subfloat[\centering][Level-based Foraging]{{
    \includegraphics[width=0.2\textwidth,trim=0 0 0 0, valign=c]{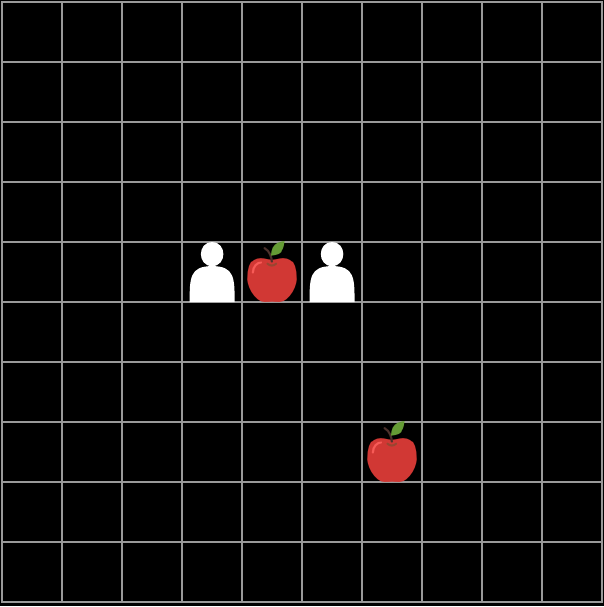}}}
    \hfill
    \subfloat[\centering][Bottleneck (4 Agents)]{{
    \includegraphics[width=0.2\textwidth,trim=0 0 0 0,valign=c]{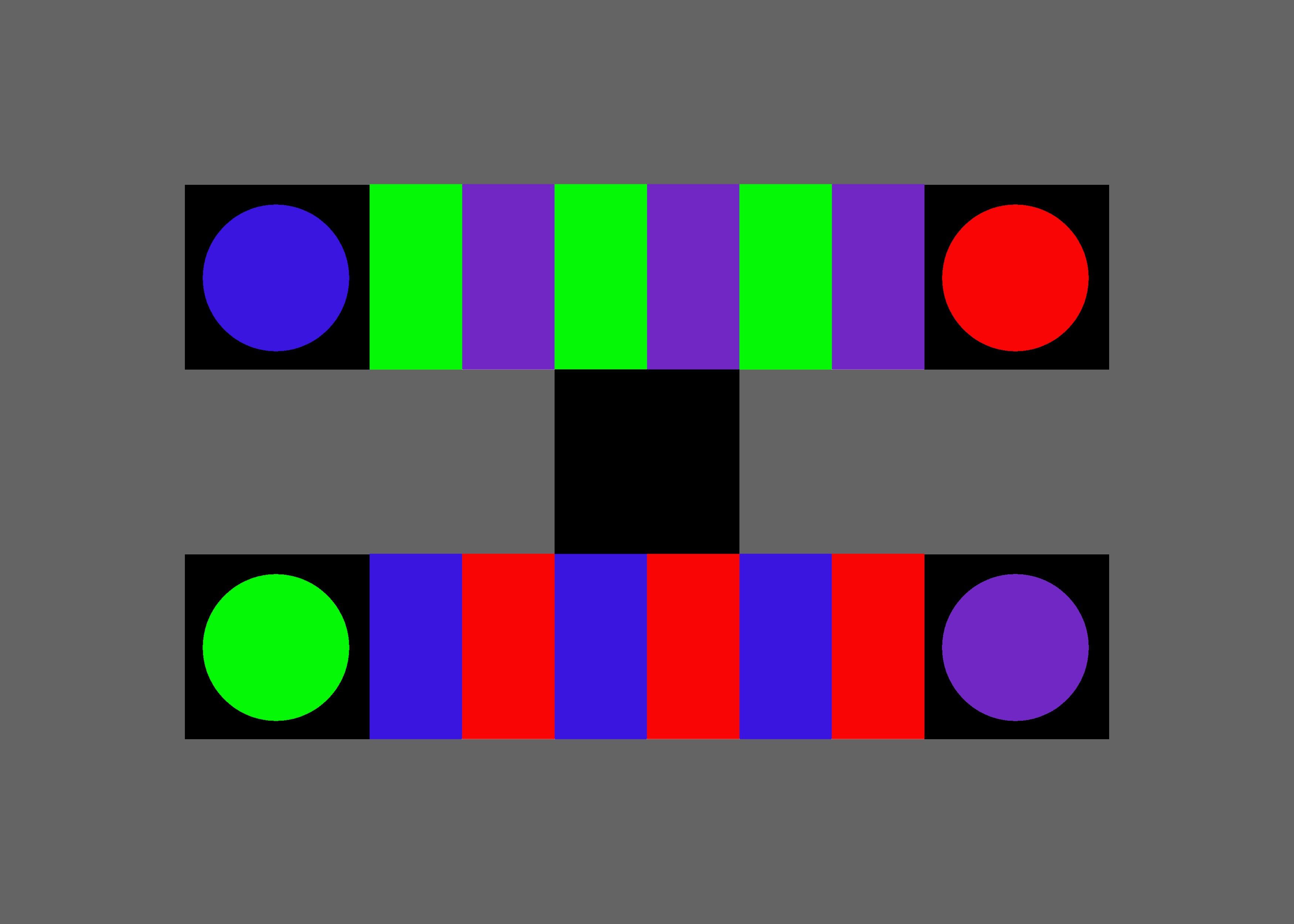}
    }}
    \hfill
    \subfloat[\centering][ClosedRooms]{{
    \includegraphics[width=0.2\textwidth,trim=0 0 0 0,valign=c]{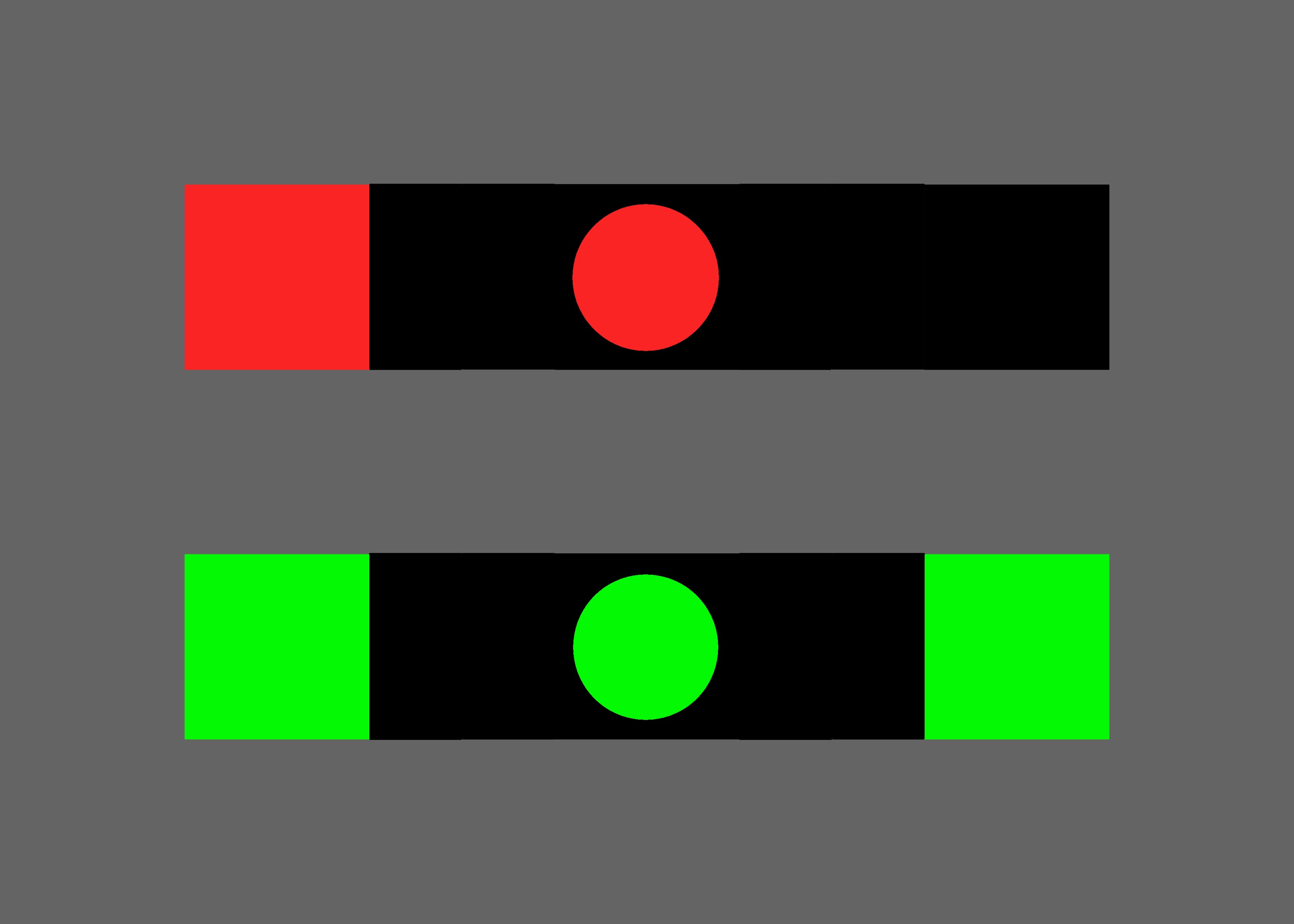}
    }}%
      \caption{Visual depiction of the different domains used in this work.}
      \label{fig:envs}%
\end{figure*}

\subsection{Variants}
We propose two additional extensions:
\begin{enumerate}[i)]
\item \textbf{Spherical ClusterComm}:
This variant investigates the effects of vector normalization. We normalize the representations using ScaleNorm~\cite{nguyen2019transformers} where they are projected onto the $(d - 1)$-dimensional hypersphere with learned radius $r$ ($r * \frac{x}{\Vert x \Vert}$).
Thus, the maximum distance is determined by $r$, which prevents individual dimensions from having to much influence and makes the features more distinctive~\cite{wang2017normface,zhai19}. In addition, the use of normalized vectors will result in clustering based on the angle between vectors~\cite{hornik2012spherical} rather than the euclidean distance.
\item \textbf{CentroidComm}:
Although one goal is to reduce the message size and use only discrete messages, we propose another alternative without these limitations. Instead of the cluster index, CentroidComm transmits the closest centroid directly. The messages thus consist of larger, continuous vectors. CentroidComm therefore uses more bandwidth for the transmission of messages during training. At the end of the training, the centroids of all other agents can be transmitted once to each agent and consequently cluster indices can be sent again at test time.
\end{enumerate}
\section{EXPERIMENTS} \label{sec:experimental-setup}
We now present the environments used to evaluate ClusterComm's performance followed by a brief discussion of the baselines and the training details.

\begin{figure*}[t]
    \centering
    \subfloat[\centering][\texttt{Bottleneck (2 Agents)}]{{\includegraphics[width=0.3\textwidth,trim=0 0 0 0]{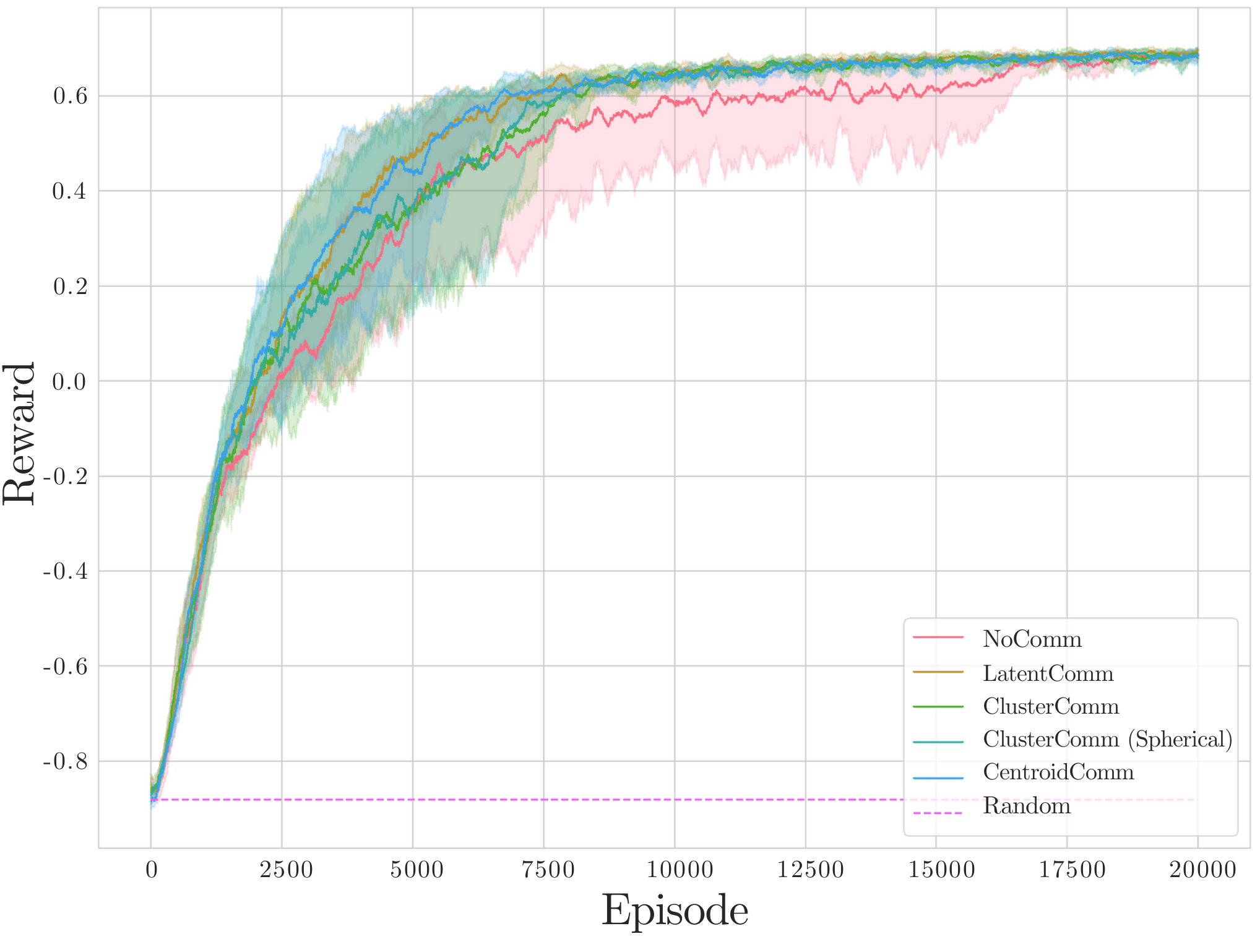} }}
    \hfill
    \subfloat[\centering][\texttt{Bottleneck (3 Agents)}]{{\includegraphics[width=0.3\textwidth,trim=0 0 0 0]{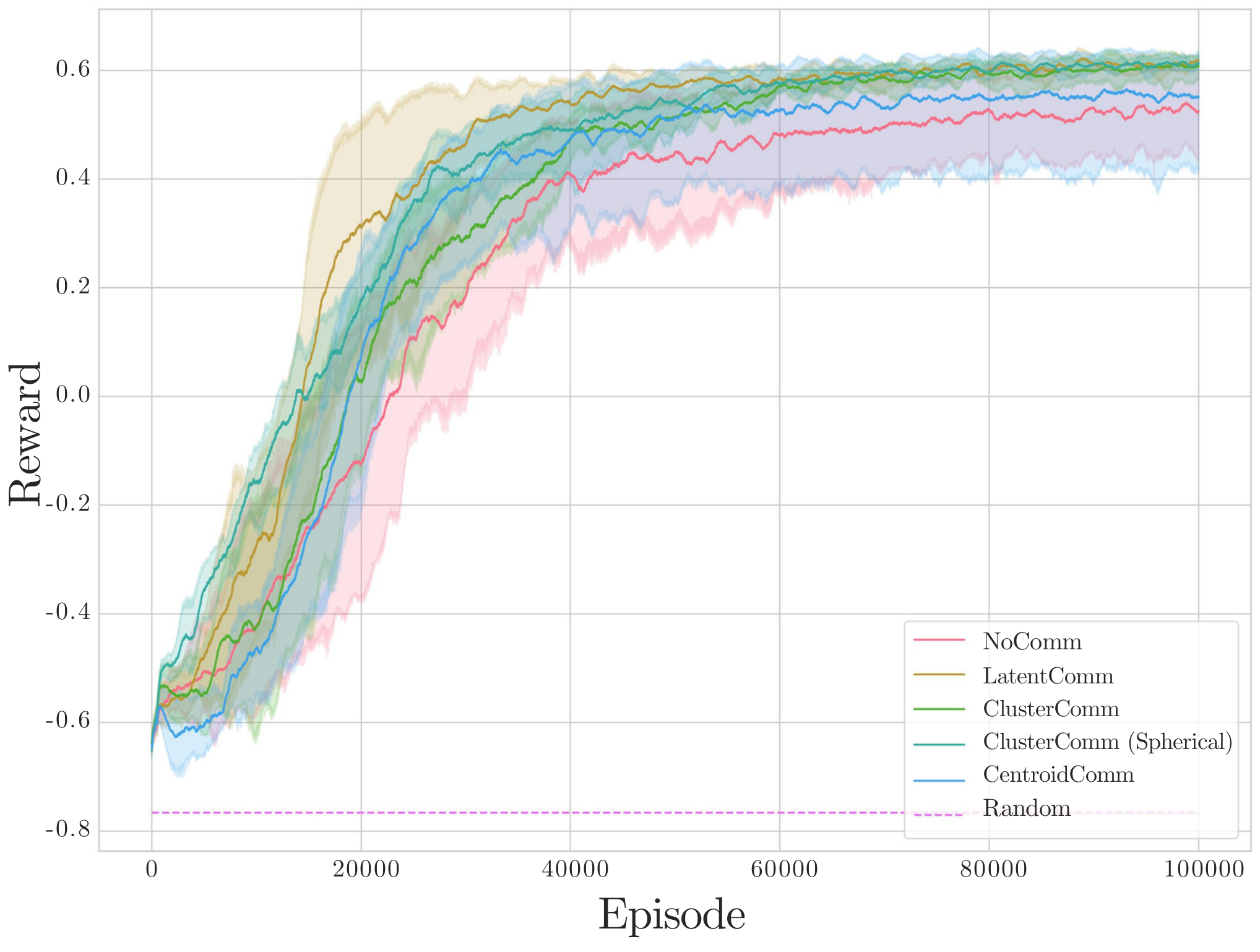} }}
    \hfill
    \subfloat[\centering][\texttt{Bottleneck (4 Agents)}]{{\includegraphics[width=0.3\textwidth,trim=0 0 0 0]{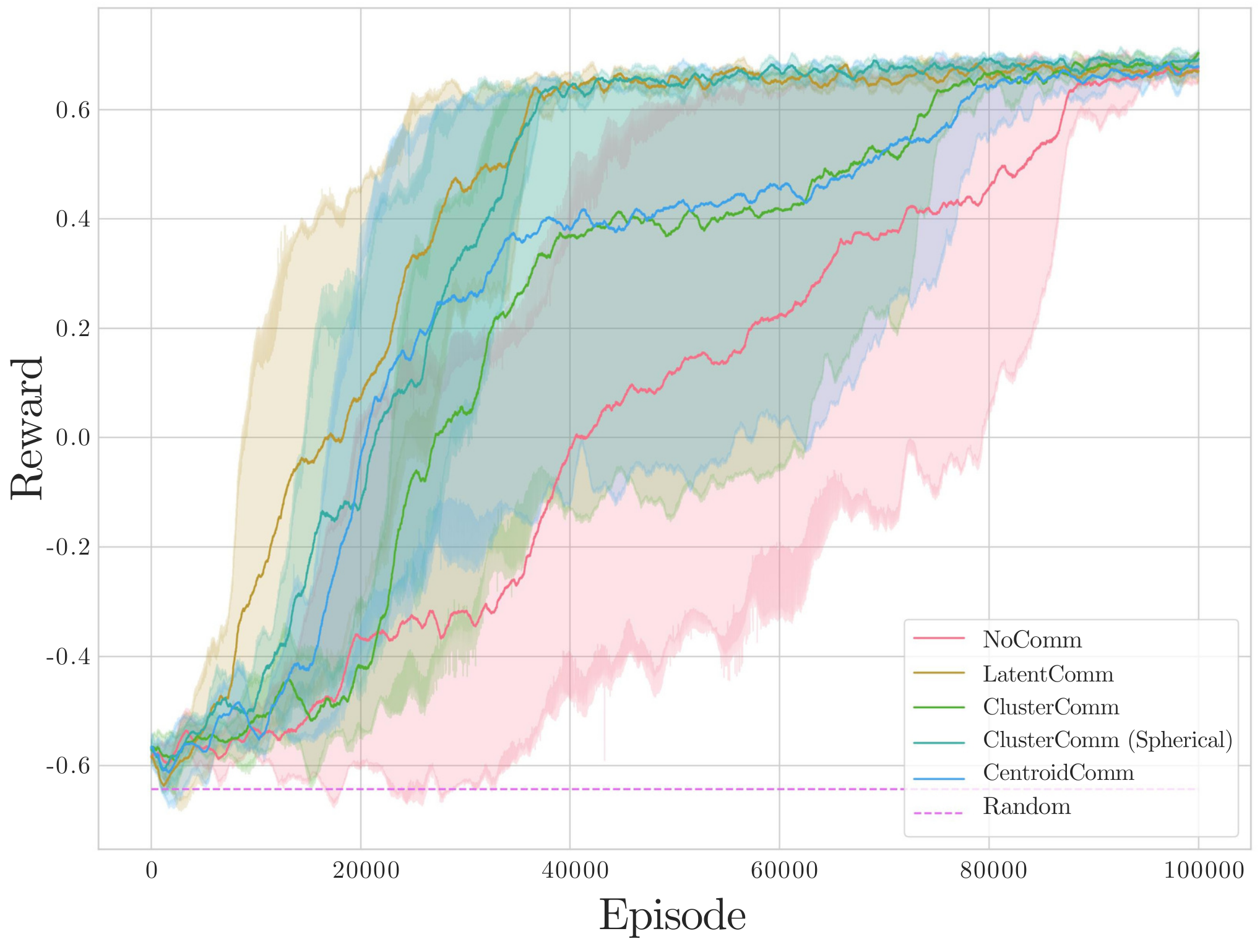} }}
    \hfill
    \subfloat[\centering][\texttt{ClosedRooms}]{{\includegraphics[width=0.3\textwidth,trim=0 0 0 0]{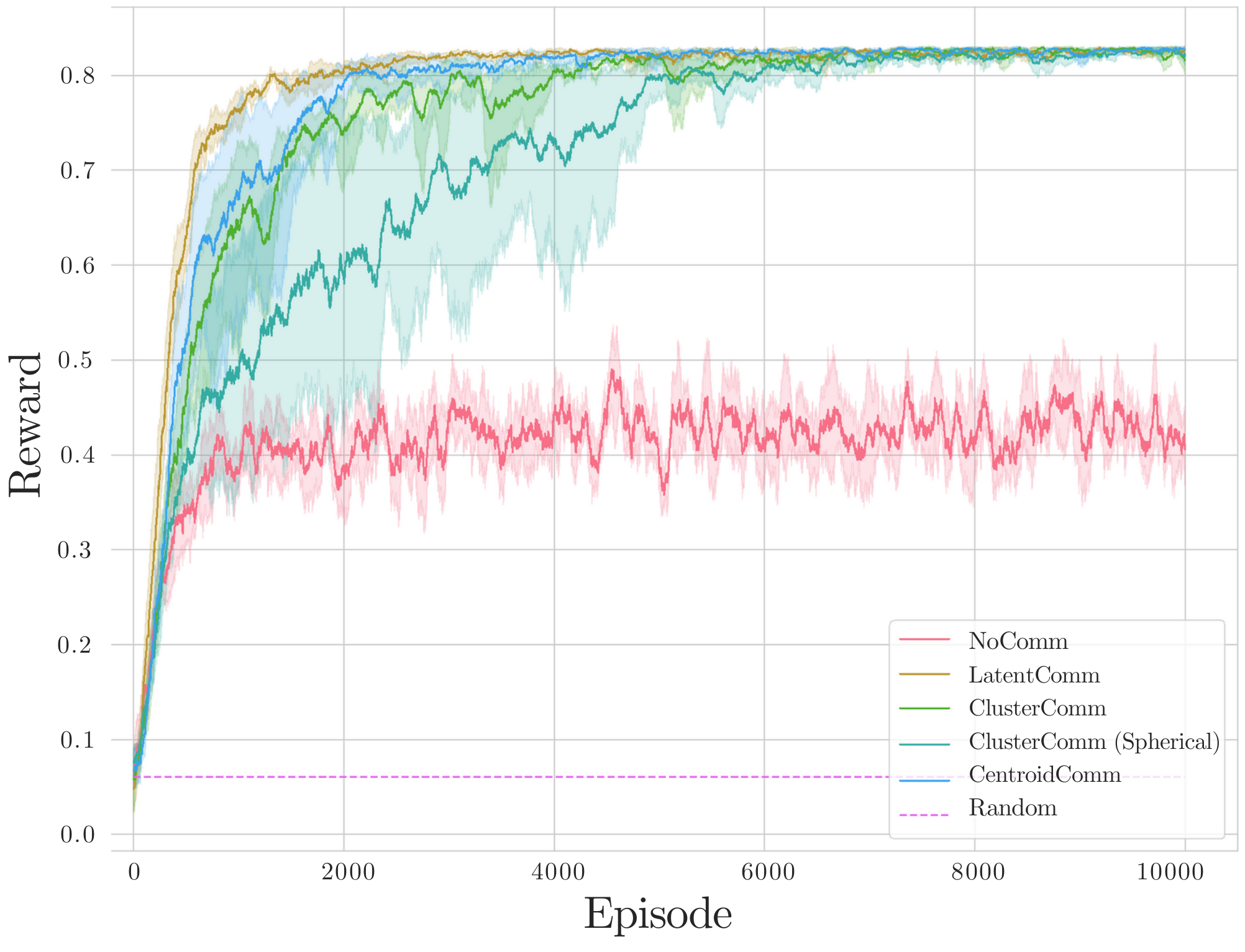} }}
    \hfill
    \subfloat[\centering][\texttt{Level-based Foraging}]{{\includegraphics[width=0.3\textwidth,trim=0 0 0 0]{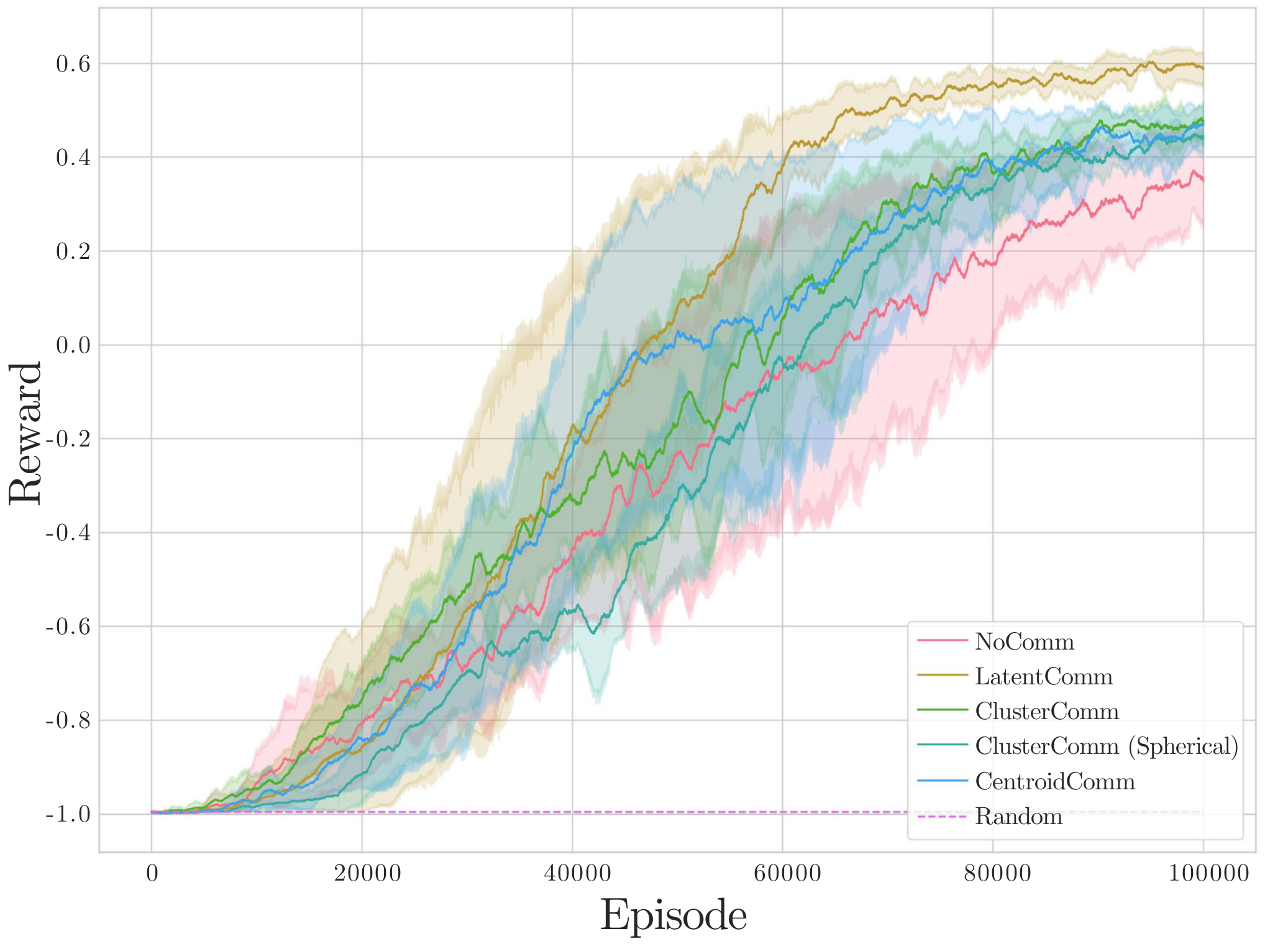} }}
    \hfill
    \subfloat[\centering][\texttt{RedBlueDoors}]{{\includegraphics[width=0.3\textwidth,trim=0 0 0 0]{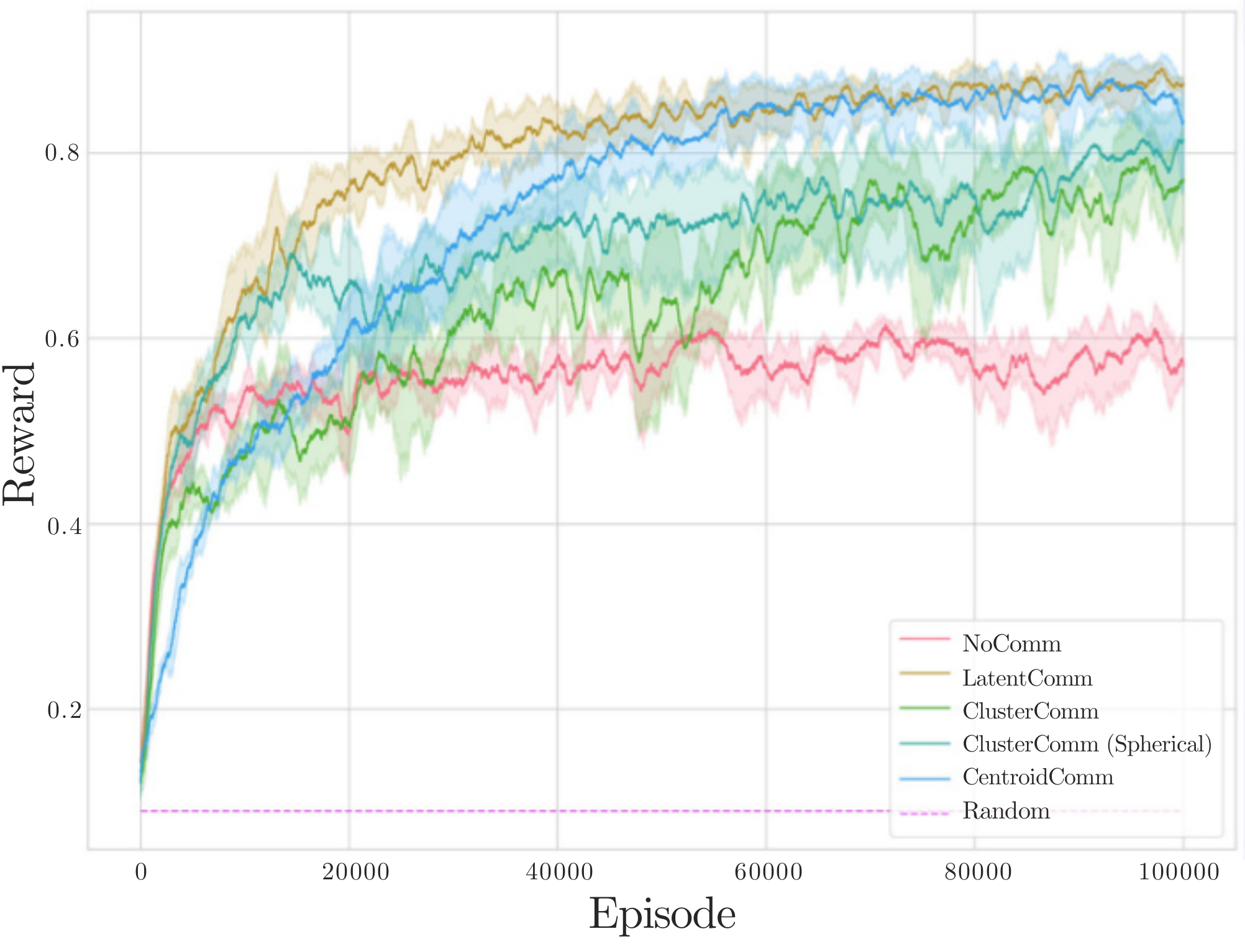} }}
   \caption{Training curves for all environments.}
   \label{fig:results}
\end{figure*}

\subsection{Environments}
The following episodic environments were used to asses ClusterComm's efficacy:
\begin{enumerate}[i)]
    \item \textbf{\texttt{Bottleneck}}:  Two rooms are connected by a single cell (bottleneck). Only one agent is allowed to occupy one cell at any time. Each agent's goal is to pass the bottleneck and reach the other side as quickly as possible. Since only one agent can move through the bottleneck at a time, the agents must coordinate their movements. 
    \item \textbf{\texttt{ClosedRooms}}: Two agents are placed in the middle of two separate, enclosed rooms. The first agent is called the speaker and the second is called listener. A corner in the speaker's room is randomly chosen and marked as the speaker's target cell. The listener's target cell is then the opposite corner in its respective room. However, the listener cannot see the speaker nor its target and must therefore rely on the speaker to communicate the goal. 
    \item \textbf{\texttt{RedBlueDoors}}~\cite{lin2021learning}:Two agents are randomly placed in an empty room. A red and a blue door are randomly positioned on the left and the right side of the room, respectively. The goal is to quickly open the red door first and the blue door afterwards. Note that since both agents can open both doors, it is possible for one agent to solve the environment on its own. 
    \item \textbf{\texttt{Level-based Foraging}}~\cite{christianos2020shared}:
    Two apples are distributed randomly in the environment. Apples can only be collected if both agents are standing on a field adjacent to the apple and decide to collect it at the same time.
\end{enumerate}
All all cases an agent's field of view was limited to $5 \times 5$ and agents do not have the ability to see through walls.
We introduce a small penalty for each step such that the agents are encouraged to solve the environments as quickly as possible.\\
Exemplary illustrations for each environment are shown in Figure~\ref{fig:envs}.
\begin{table*}[t]
{\tiny
\begin{center}
\caption{Performance of trained agents in the \texttt{Bottleneck} environment.}
\label{tab:bottleneck}
\begin{tabularx}{\linewidth}{Xccccccccc}
\toprule
\multicolumn{10}{c}{\texttt{\textbf{Bottleneck}}} \\
\midrule
\multicolumn{1}{c}{} & \multicolumn{3}{c}{2 Agents}  & \multicolumn{3}{c}{3 Agents} & \multicolumn{3}{c}{4 Agents} \\
\midrule
Algorithm & $\varnothing$Rew. &$\varnothing$ Succ. & $\varnothing$Steps & $\varnothing$Rew. &$\varnothing$ Succ. & $\varnothing$Steps &$\varnothing$Rew. &$\varnothing$ Succ. & $\varnothing$Steps \\ 
\midrule
NoComm & 0.70 & 0.99 & 9.17 & 0.52 & 0.93 & 12.84 & 0.67 & 0.97 & 18.73 \\ 
LatentComm & 0.70 & 0.99 & 9.22 & 0.60 & 0.98 & 12.07 & 0.67 & 0.97 & 19.04 \\ 
ClusterComm & 0.70 & 0.99 & 9.39 & 0.60 & 0.98 & 12.21 & 0.70 & 0.98 & 17.78 \\
ClusterComm (Spher.) & 0.71 & 0.99 & 9.14 & 0.60 & 0.98 & 12.08 & 0.69 & 0.97 & 17.83 \\
CentroidComm & 0.71 & 0.99 & 9.11 & 0.55 & 0.92 & 11.93 & 0.67 & 0.96 & 18.45 \\
Random & -0.87 & 0.008 & 26.11 & -0.76 & 0.0 & 26.83 & -0.64 & 0.0 & 55.37 \\ 
\bottomrule
\end{tabularx}
\end{center}
}
\end{table*}

\begin{table*}[t]
{\tiny
\begin{center}
\caption{Performance of trained agents in the \texttt{ClosedRooms}, \texttt{Level-based Foraging} and \texttt{RedBlueDoors} environments.}
\label{tab:rest}
\begin{tabularx}{\linewidth}{Xccccccccc}
\toprule
\multicolumn{1}{c}{} & \multicolumn{3}{c}{\texttt{\textbf{ClosedRooms}}}  & \multicolumn{3}{c}{\texttt{\textbf{Level-based Foraging}}} & \multicolumn{3}{c}{\texttt{\textbf{RedBlueDoors}}} \\
\midrule
Algorithm & $\varnothing$Rew. &$\varnothing$ Succ. & $\varnothing$Steps & $\varnothing$Rew. &$\varnothing$ Succ. & $\varnothing$Steps &$\varnothing$Rew. &$\varnothing$ Succ. & $\varnothing$Steps\\ 
\midrule
NoComm &        0.40 & 0.49 & 2.60 & 0.39 & 0.87 & 60.29 & 0.58 & 0.63 & 48.09 \\ 
LatentComm &    0.82 & 0.99 & 3.05 & 0.60 & 0.98 & 47.52 & 0.86 & 0.91 & 34.20 \\ 
ClusterComm &   0.82 & 0.99 & 3.05 & 0.47 & 0.92 & 55.53 & 0.70 & 0.75 & 39.92 \\
ClusterComm (Spher.) &  0.82 & 0.99 & 3.05 & 0.46 & 0.93 & 57.49 & 0.77 & 0.83 & 36.32 \\
CentroidComm &  0.82 & 0.99 & 3.04 & 0.42 & 0.88 & 57.58 & 0.84 & 0.88 & 26.00 \\
Random &        0.06 & 0.12 & 9.18 & -0.99 & 0.00& 128.00 & 0.08 & 0.16 & 272.44 \\ 
\bottomrule
\end{tabularx}
\end{center}
}
\end{table*}

\subsection{Baselines and Training Details}
We compare ClusterComm with the following three baselines:
\begin{enumerate}[i)]
    \item \textbf{Random}: Each action is selected randomly with equal probability.
    \item \textbf{NoComm}: The agents are trained without a communication channel.
    \item \textbf{LatentComm}~\cite{lin2021learning}: The entire representation is transmitted. This approach can be regarded as unrestricted, since the agents transmit continuous, high-dimensional vectors. 
\end{enumerate}
The goal is to show that ClusterComm achieves similar performance to LatentComm, even though the agents are only allowed to transmit discrete messages. Although useful information may be lost through the use of clustering, ClusterComm is more scalable and reduces the required bandwidth.\\
For ClusterComm, its variations, as well as the aforementioned baselines, an MLP with the Tanh activation function and two hidden layers of size $32$ is used for the message and observation encoder. Both outputs are concatenated and a linear layer is used to predict the distribution over actions. Moreover, we use frame-stacking with the last three observations. ClusterComm updates the centroids after each PPO update. The number of clusters is $16$ for RedBlueDoors and Level-based Foraging and $8$ for Bottleneck and ClosedRooms.\\
To evaluate the performance of trained agents, we employ the trained policies over $1000$ additional episodes and record the average reward, the average number of steps and the success rate (Table~\ref{tab:bottleneck} and Table~\ref{tab:rest}). All experiments are repeated $10$ times.

\section{RESULTS} \label{sec:results}
In this section, we discuss the results derived from Figure~\ref{fig:results}, Table~\ref{tab:bottleneck} and Table \ref{tab:rest}. For clarity, these are discussed separately for each environment.

\subsection{Bottleneck}
 The results are shown in Figures~\ref{fig:results} a) to c) and Table~\ref{tab:bottleneck}. 
 In these experiments, the scalability of ClusterComm in terms of the number of agents is of particular interest. The more agents are included, the more difficult the task becomes. \\
One can see that the difference between the approaches becomes more obvious as the complexity increases. With two agents (Figure~\ref{fig:results} a)), all methods perform equally well and converge to the optimal solution. However, LatentComm was found to have the fastest convergence, while NoComm takes more time to converge and exhibits a less stable learning curve. The results for three agents are similar. Here, the difference is that the methods need more episodes to converge and NoComm fails to find an equally effective policy in this case. With four agents (figures~\ref{fig:results} c)), the differences between the approaches increase further. LatentComm still achieves the best results, followed by CentroidComm, ClusterComm and finally NoComm. In addition, a greater variance than before can be observed. Due to the considerably increased coordination effort, the agents have to explore significantly more during training in order to finally find a suitable policy. Here, all ClusterComm approaches help to find it earlier and more reliably.

\subsection{ClosedRooms}
The results are shown in Figure~\ref{fig:results} d). Since this environment requires communication, methods like Random or NoComm cannot solve the task. NoComm converges to the optimal strategy possible without communication, i.e. the listener chooses a random target in each episode. Thus, the agents successfully solve the problem in 50\% of all cases.\\
Approaches with communication learn to solve the problem completely. However, they differ in learning speed and stability. It is evident that LatentComm converges faster and also exhibits lower variance. ClusterComm and CentroidComm learn slightly slower, but also converge to the same result. Furthermore, the variance of ClusterComm is lower than that of CentroidComm. This can be attributed to the message type. After the cluster is updated, the cluster index often remains the same, while the centroid changes continuously. This means that the ClusterComm messages change less frequently for the same observations, while the CentroidComm messages change after each update until the procedure converges.\\
With respect to table~\ref{tab:rest}, all methods with communication yield the same optimal results (99\% success rate with an average of 3 steps per episode). NoComm successfully solves only half of the episodes, but with a slightly lower average number of steps. Agents trained with NoComm choose a corner at random and move towards it. The low number of steps in NoComm is therefore due to the fact that although agents quickly move to a random corner and terminate the episode, this is the correct solution in only half of all the cases.
In contrast, through the use of communication, the listener learns to stop at the first time step and wait for the message sent by the speaker.

\subsection{RedBlueDoors}
Once again it becomes evident (Figure~\ref{fig:results} f)) that the ability to communicate is advantageous. All methods with communication achieve very good results. LatentComm learns the fastest, closely followed by CentroidComm. Both approaches converge to an optimal value. The performance of ClusterComm is slightly worse, but still better than NoComm. \\
The results from table~\ref{tab:rest} confirm that LatentComm (success rate $91$\%) and CentroidComm (success rate $88$\%) are the strongest, followed by ClusterComm, whose success rate is $75$\% and requires more steps to complete an episode.\\
Also noticeable is that CentroidComm takes fewer steps on average than LatentComm, although LatentComm has a higher success rate. We observe that CentroidComm needs significantly fewer steps in a successful episode than LatentComm. However, agents do not coordinate sufficiently in more cases compared to LatentComm.\\ Although the continuous adaptation of the representations during the training of LatentComm leads to the agents taking longer to solve the problem, it increases their success rate. By sending the centroids, the variance of the individual features is limited, since representations of different observations can be mapped onto one and the same centroid. At the same time, this is the biggest disadvantage of CentroidComm, as information is lost, which reduces the success rate. 

\subsection{Level-based Foraging}
It is again clear that all communication-based approaches outperform NoComm, with LatentComm providing the best results (Figure~\ref{fig:results} e)) due to its continuous message type. Moreover, all ClusterComm approaches show roughly the same performance, with a slight advantage for ClusterComm (Table~\ref{tab:rest}).\\
Not only do the agents have to find the position of the food sources, but they also have to agree on which food source both agents should move to. The established communication protocol helps the agents to make the decision.\\

\noindent In general, the results suggest that none of the ClusterComm variants presented is clearly superior to the others. Their performance strongly depends on the environment, the task to be solved and the resulting training dynamics.

\section{CONCLUSION} \label{sec:conclusion}
In this work, we introduced ClusterComm, a MARL algorithm aimed at enhancing communication efficiency among agents. ClusterComm discretizes the internal representations from each agent's observations and uses the resulting cluster indices as messages. Unlike prevalent centralized learning methods, ClusterComm fosters independent learning without parameter sharing among agents, requiring only a discrete communication channel for message exchange. Our empirical evaluations across diverse environments consistently demonstrated ClusterComm's superiority over the no-communication approach and yielded competitive performance to LatentComm which uses unbounded communication. Future work may involve reducing the dependence of the network architecture on the number of agents, exploring multi-index message transmission and extending the communication phase to address information loss, resolve ambiguities or facilitate negotiation.

\section*{ACKNOWLEDGEMENTS}
This work was funded by the Bavarian Ministry for Economic Affairs, Regional Development and Energy as part of a project to support the thematic development of the Institute for Cognitive Systems.

\bibliographystyle{apalike}
{\small
\bibliography{main}}

\end{document}